\documentclass[10pt,twocolumn,letterpaper]{article}

\usepackage{cvpr}
\usepackage{times}
\usepackage{epsfig}
\usepackage{graphicx}
\usepackage{amsmath}
\usepackage{float}
\usepackage{amssymb}
\usepackage{multirow}
\newcommand\blfootnote[1]{%
  \begingroup
  \renewcommand\thefootnote{}\footnote{#1}%
  \addtocounter{footnote}{-1}%
  \endgroup
}

\usepackage[pagebackref=true,breaklinks=true,letterpaper=true,colorlinks,bookmarks=false]{hyperref}

\cvprfinalcopy 

\def\cvprPaperID{****} 

\begin{document}

\title{ NITS-VC System for VATEX Video Captioning Challenge 2020}

 \author{\textbf{Alok Singh}, \textbf{Thoudam Doren Singh} and \textbf{Sivaji Bandyopadhyay}\\
 Center for Natural Language Processing \& Department of Computer Science and Engineering\\
 National Institute of Technology Silchar\\
 Assam, India\\
 {\tt\small \{alok.rawat478, thoudam.doren, sivaji.cse.ju\}.gmail.com}
}

\maketitle

\begin{abstract}

 Video captioning is process of summarising the content, event and action of the video into a short textual form which can be helpful in many research areas such as video guided machine translation, video sentiment analysis and providing aid to needy individual. In this paper, a system description of the framework used for VATEX-2020 video captioning challenge is presented. We employ an encoder-decoder based approach in which the visual features of the video are encoded using 3D convolutional neural network (C3D) and in the decoding phase two Long Short Term Memory (LSTM) recurrent networks are used in which visual features and input captions are fused separately and final output is generated by performing element-wise product between the output of both LSTMs. Our model is able to achieve BLEU scores of 0.20 and 0.22 on public and private test data sets respectively.  
 
\end{abstract}


\section{Introduction}

Nowadays, the amount of multimedia data (especially video) over the Internet is increasing day by day. This leads to a problem of automatic classification, indexing and retrieval of the video \cite{singh2019novel}. Video captioning is a task of automatic captioning a video by understanding the action and event in the video which can help in the retrieval of the video efficiently through text. On addressing the task of video captioning effectively, the gap between computer vision and natural language can also be minimized. Based on the approaches proposed for video captioning till now, they can be classified into two categories namely, template-based language model \cite{barbu2012video} and sequence learning model \cite{pan2016hierarchical}. The template based approaches use predefined templates for generating the captions by fitting the attributes identified in the video. These kinds of approaches need the proper alignment between the words generated for the video and the predefined templates. In contrast to template based approach, the sequence learning based approach learn the sequence of word conditioned on previously generated word and visual feature vector of the video. This approach is commonly used in Machine Translation (MT) where the target language (T) is conditioned on the source language (S).

Video is a rich source of information which consists of large number of continuous frames, sound and motion. The presence of large number of similar frames, complex actions and events of video makes the task of video captioning challenging. Till now a number of data sets have been introduced by covering variety of domains such as cooking \cite{Das_2013_CVPR}, social media \cite{gella2018dataset} and human action \cite{chen2011collecting}. Based on the data sets available for video captioning, the approaches proposed for video captioning can be categorized into open domain video captioning \cite{wang2019vatex} and  domain specific video captioning \cite{Das_2013_CVPR}. The process of generating caption for an open domain video is challenging due to the presence of various intra-related action, event and scenes as compared to domain specific video captioning. In this paper, an encoder-decoder based approach is used for addressing the problem of English video captioning in a dataset provided by the VATEX community for multilingual video captioning challenge 2020 \cite{wang2019vatex}. 

\section{Background Knowledge}
The background of video captioning approaches can be divided into three phases \cite{aafaq2019video}. The classical video captioning approach phase involves the detection of entities of the video (such as object, actions and scenes) and then map them to a predefined templates. The statistical methods phase, in which the video captioning problem is addressed by employing statistical methods. The last one is deep learning phase. In this phase, many state-of-the-art video captioning framework have been proposed and it is believed that this phase has a capability of solving the problem of automatic open domain video captioning.

Some classical video captioning approaches are proposed in \cite{brand1997inverse, koller1991algorithmic} where motion of vehicles in a video  and series of actions in a movie are characterized using natural language respectively. The successful implementation of Deep Learning (DL) techniques in the field of computer vision and natural language processing attract researchers for incorporating DL based techniques for text generation. Most of these DL based approaches for video captioning are inspired by the deep image captioning approaches \cite{xu2019semantic}. These DL based approaches for video captioning mainly consist of two stages namely, visual feature encoding stage and sequence decoding stage.  \cite{venugopalan2015sequence} proposed a model based on stacked LSTM in which first LSTM layer takes a sequence of frames as an input and second LSTM layer generates the description using first LSTM hidden state representation. The shortcoming of the approach proposed in \cite{venugopalan2015sequence} is that all frames need to be processed for generating description and since the video consist of many similar frames, there is a chance that final representation of the visual features consist of less relevant information for video captioning \cite{xu2019semantic}. Some other DL based technique are successfully implemented in  \cite{donahue2015long, venugopalan2014translating}. In this paper, the proposed architecture and performance of the system is presented which is evaluated on VATEX video captioning challenge dataset 2020\footnote{\url{https://competitions.codalab.org/competitions/24360}}. Furthermore, in Section \ref{sec:mep}, description of system used for challenge is given, followed by Section \ref{sec:ds} and Section \ref{sec:con} on discussion of performance of the system and conclusion respectively.

\section{VATEX Task} \label{sec:mep}
In this section, detail of visual feature extraction of the video and the implementation of the system are discussed. 

\subsection{System Description} \label{sec:modl}
\textbf{Visual Feature Encoding:} For this task, a traditional encoder-decoder based approach is used. For encoding, the visual feature vector of the input video, firstly, the video is evenly segmented into $n$ segments in the interval of $16$ then, using a  pre-trained 3D Convolutional Neural Network (C3D), a visual feature vector $f=\{s_1, s_2\dots s_n\}$ for video is extracted. The dimension of feature vector for each video is $f\in \mathbb{R}^{n\times m_x}$ where $m_x$ is dimension feature vector for each segment. Since, the high dimensional feature vector is prone to carry redundant information and affect the quality of features \cite{xu2019semantic}, for reducing the dimension of features an average pooling is performed with filter size $5$. All the averaged pooled features($f_a$) of each segment are concatenated in a sequence to preserve the temporal relation between them and passed to a decoder for caption generation.

\textbf{Caption generator:} For the decoding, two separate Long Short Term Memory (LSTM) recurrent networks are used as a decoder. In this stage, firstly, all the input captions are passed to a embedding layer to get a dense representation for each word in the input caption. The embedding representation is then passed to both LSTM separately. The first LSTM takes the encoded visual feature vector as an initial stage and for the second LSTM, the visual feature vector concatenated with the output of embedding layer and finally the element wise product is preformed between the output from both LSTM. The unrolling procedure of system is given below:
\blfootnote{[ ; ] represent the concatenation}
\begin{equation}\label{eq:1}
    \tilde{y} = W_eX +b_e 
\end{equation}
\begin{equation}\label{eq:2}
    \tilde{z_1} = LSTM_{1}(\tilde{y},h_i)
\end{equation}
\begin{equation}\label{eq:3}
    \tilde{z_2} = LSTM_{2}([\tilde{y};f_a])
\end{equation}
\begin{equation}\label{eq:4}
    y_t = softmax(\tilde{z_1}\odot\tilde{z_2})
\end{equation}
The Equation \ref{eq:1} represents the embedding of input captions ($X$) where $W_e$ and $b_e$ are learnable parameter weight and bias respectively. The $\tilde{z_1}$ in Equation \ref{eq:2} and $\tilde{z_2}$  in Equation \ref{eq:3} are the output of LSTM layers where, $h_i$ ($h_i = f_a$) is the average pooled feature vector which is passed as a hidden state for $LSTM_1$ and for $LSTM_2$ it concatenated with embedding vector. $\odot$ represent the product of $\tilde{z_1}$ and $\tilde{z_2}$ which is finally passed to softmax layer. 

\begin{table}[!ht]\scriptsize
    \caption{Statistics of dataset used}
    \label{tab:datasetd}
    \centering
    \begin{tabular}{c|c|c|c}
    \hline
    {\textbf{Dataset }} &\multirow{2}{*}{\textbf{ \#Videos}} &  {\textbf{\#English}} &{\textbf{\#Chinese}}\\ 
        \textbf{Split}&&\textbf{Captions}&\textbf{Captions}\\ \hline
          Training& 25,991 & 259,910 &259,910\\
         Validation& 3,000 & 30,000 &259,910\\
         Public test set& 6,000 &30,000 &30,000\\
         Private test set&6,287 &62,780 &62,780 \\\hline
    \end{tabular}
\end{table}

\begin{table*}[!ht]\scriptsize
   \caption{Sample output caption generated by system.}
     \label{tab:op_cap}
    \centering
    \begin{tabular}{p{2cm} p{2cm}| p{2cm} p{2cm}| p{2cm} p{2cm}}
       \includegraphics[width=2cm,height=2cm]{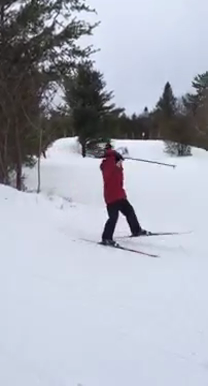}&\includegraphics[width=2cm,height=2cm]{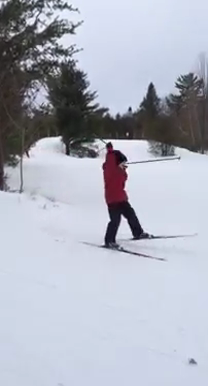} &\includegraphics[width=2cm,height=2cm]{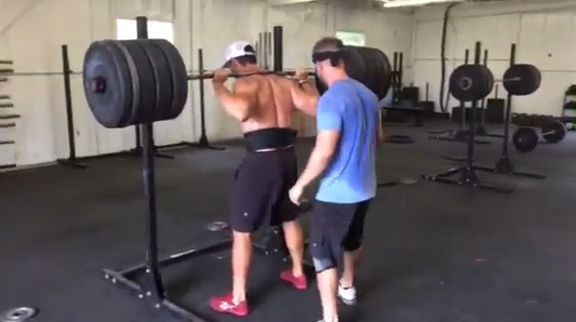} &\includegraphics[width=2cm,height=2cm]{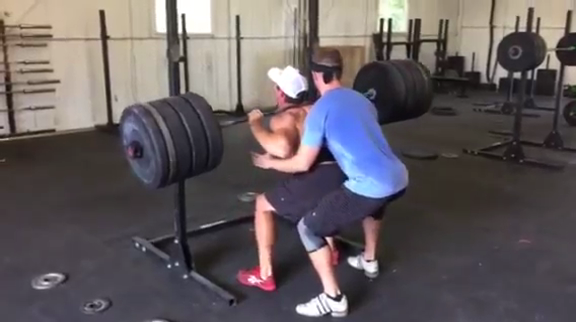}&\includegraphics[width=2cm,height=2cm]{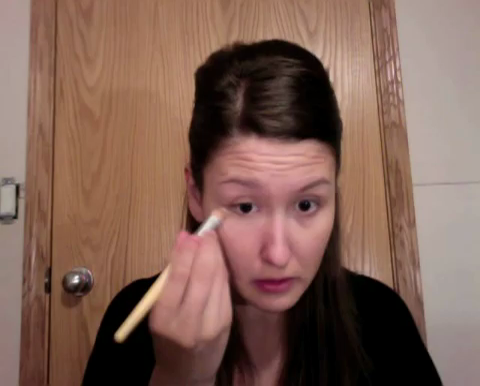} &\includegraphics[width=2cm,height=2cm]{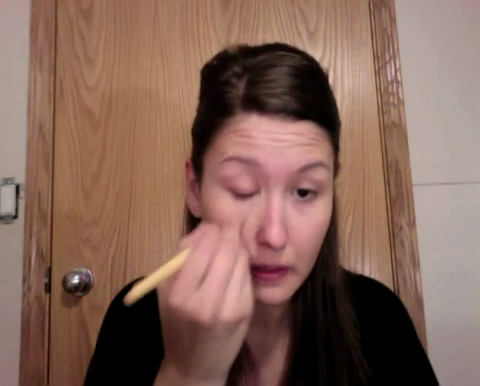}\\
        \includegraphics[width=2cm,height=2cm]{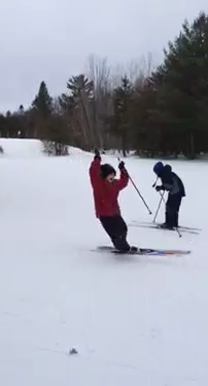}&\includegraphics[width=2cm,height=2cm]{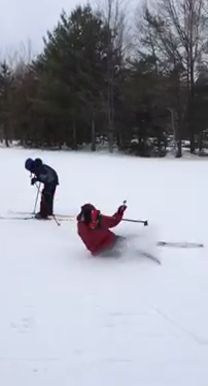} &\includegraphics[width=2cm,height=2cm]{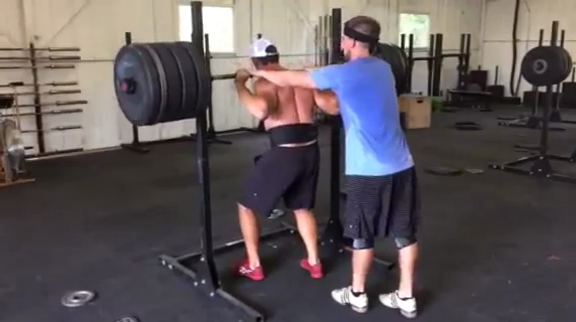} 
         &\includegraphics[width=2cm,height=2cm]{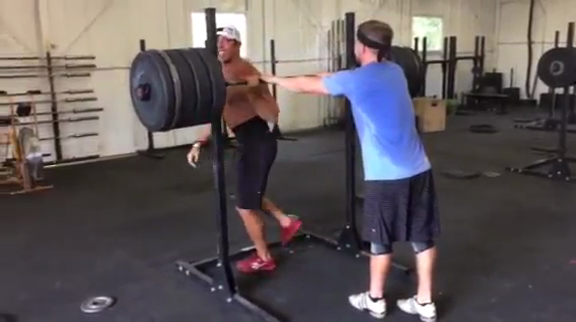}&\includegraphics[width=2cm,height=2cm]{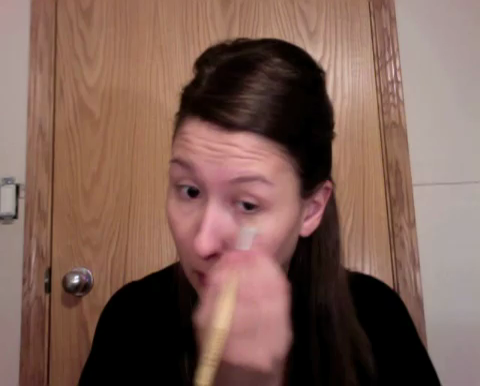} &\includegraphics[width=2cm,height=2cm]{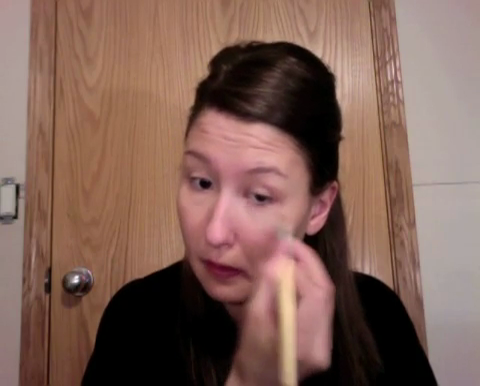}\\
\multicolumn{2}{c|}{\textbf{VideoId-2-40BG6NPaY}}&\multicolumn{2}{c|}{\textbf{VideoId-fjErVZXd9e0}}&\multicolumn{2}{c}{\textbf{VideoId-IwjWR7VJiYY}}\\

    \multicolumn{2}{c|}{\textbf{Generated Caption}}&\multicolumn{2}{c|}{\textbf{Generated Caption}}&\multicolumn{2}{c}{\textbf{Generated Caption}}\\

  \multicolumn{2}{p{4.5cm}|}{A group of people are skiing down a hill and one of them falls down}& \multicolumn{2}{p{4.5cm}|}{A man is lifting a heavy weight over his head and then drops it}&\multicolumn{2}{p{4.5cm}}{A woman is demonstrating how to apply mascara to her eyelashes}\\

  \multicolumn{2}{c|}{\textbf{(a)}}&\multicolumn{2}{c|}{\textbf{(b)}}&\multicolumn{2}{c}{\textbf{(c)}}\\
    \end{tabular}
\end{table*}

\section{Experimental Results and Discussion} \label{sec:ds}
In this section, a brief discussion on dataset used, experimental setup and the output generated by the system are carried out.


\subsection{Dataset}
For the evaluation of the system performance, VATEX video captioning dataset is used. This dataset is proposed for motivating multilingual video captioning, in which each video is associated with corresponding $10$ captions in both English and Chinese. This dataset consist of two test sets namely, public test set for which reference captions are publicly available and private test set for which reference captions are not publicly available which are on hold for the evaluation in VATEX video captioning challenge 2020. The detailed statistics of the dataset is given in Table \ref{tab:datasetd}. The system employed in the paper is implemented using QuADro P2000 GPU and tested on both the test sets.


\subsection{Evaluation Metrics}
For the evaluation of generated captions, various evaluation metrics have been proposed. In this paper, for the quantitative evaluation of the generated output, the following evaluation metrics are used: BLEU\cite{papineni2002bleu}\footnote{\url{https://github.com/tylin/coco-caption}}, METEOR \cite{denkowski2014meteor}, CIDEr \cite{vedantam2015cider} and ROUGE-L. BLEU evaluates the part of n-grams (up to four) that are similar in both reference (or a set of references) and hypothesis. CIDEr also measure the n-grams that are common in both hypothesis and the references, but in CIDEr, term frequency-inverse document is used for weighting the n-grams. In METEOR, once the common unigrams are found, then it calculates a score for this matching using a unigram-precision, unigram-recall and measure of fragmentation which is used for evaluating how well-ordered the matched word in generated caption against the reference caption. 
\subsection{Experimental Setup}
Following the system architecture discussed in Section \ref{sec:modl}, after extracting the spatio-temporal features using C3D model pre-trained on Sports-1M dataset \cite{karpathy2014large,tran2015learning}, they are passed to caption generator module.

In the training process, each caption is concatenated by two special marker $<BOS>$ and $<EOS>$ for informing the model about the beginning and ending of caption generation process. We restricted the maximum number of words in a caption to $30$, and if the length of the caption is less than desired length the caption is padded with $0$. The captions are tokenized using Stanford tokenizer \cite{manning-EtAl:2014:P14-5} and only to $15K$ words with most occurrence are retained. For out-of-vocabulary words, a special tag $UKN$ is used. In the testing phase, the generation of caption starts after watching start marker and the input visual feature vector, in each iteration, the most probable word is sampled out and passed to next iteration with previous generated words and visual feature vector until a special end marker is generated. For loss minimization, cross-entropy loss function is used with $ADAM$ optimizer and learning rate is set to $2\times10^{-4}$. For reducing the overfitting situation, a dropout of $0.5$ is used and the hidden units of both LSTMs are set to $512$.  The system is trained with different batch size $64$ and $256$ for 50 epochs each. On analysing the scores of evaluation metrics and the quality of generated captions in terms of coherence and relevancy, smaller batch size performs better.

\subsection{Results}
The performance of the system on private and public test set is given in Table \ref{tab:pbper}.
\begin{table}[H]\scriptsize
 \caption{Performance of the system on public dataset}
    \label{tab:pbper}
    \centering
    \begin{tabular}{c|c|c}
    \hline
    {\textbf{Evaluation}} &{\textbf{Proposed System}}&{\textbf{Proposed System}} \\
      \textbf{Metrics} &\textbf{on public test set}&\textbf{on private test set}\\\hline
        CIDEr  & 0.24  &0.27 \\
        BLEU-1 & 0.63  &0.65 \\
        BLEU-2 & 0.43 &0.45 \\
        BLEU-3 & 0.30 &0.32 \\
        BLEU-4 & 0.20 &0.22  \\
        METEOR & 0.18  &0.18 \\
        ROUGE-L &0.42 &0.43 \\\hline
    \end{tabular}
\end{table}

In the Table \ref{tab:op_cap}, sample output captions generated by the system is given along with videoId.

\section{Conclusion}\label{sec:con}
In this paper, a description of the system which is used for VATEX2020 video captioning challenge is presented. We have used encode-decoder based video captioning framework for the generation of English captions. Our system scored $0.20$ and $0.22$ BLEU-4 score on public and private video captioning test set respectively.

\section*{Acknowledgments}
This work is supported by Scheme for Promotion of Academic and Research Collaboration (SPARC) Project Code: P995 of No: SPARC/2018-2019/119/SL(IN) under MHRD, Govt of India.

 {\small
 \bibliographystyle{ieee_fullname}
 \bibliography{egbib}
 }

\end{document}